\documentclass{article} 
\usepackage{iclr2027_conference,times}

\usepackage{amsmath,amsfonts,bm}

\def\eqref#1{equation~\ref{#1}}

\def\1{\bm{1}}

\DeclareMathAlphabet{\mathsfit}{\encodingdefault}{\sfdefault}{m}{sl}
\SetMathAlphabet{\mathsfit}{bold}{\encodingdefault}{\sfdefault}{bx}{n}

\usepackage{hyperref}
\usepackage{url}
\usepackage{graphicx}
\usepackage{subcaption}
\usepackage{wrapfig}
\usepackage{booktabs}
\usepackage{multirow}
\usepackage{siunitx}
\usepackage{amssymb} 

\usepackage[capitalize]{cleveref}

\title{Move, don't think: sample-efficient collision avoidance with lossy proximity sensors}

\author{Carson Kohlbrenner$^{1}$ \quad
Niraj Pudasaini$^{2}$ \quad
William Xie$^{2}$ \\
\textbf{Naren Sivagnanadasan}$^{1}$ \quad
\textbf{Nikolaus Correll}$^{2}$ \quad
\textbf{Alessandro Roncone}$^{1}$ \\[0.5em]
$^{1}$Human Interaction and Robotics Group \quad
$^{2}$Correll Lab \\
University of Colorado Boulder \\
\texttt{\{first.last\}@colorado.edu}
}

\iclrpreprint 
\begin{document}

\maketitle

\begin{abstract}

Data-driven robots must learn to avoid collisions with as few demonstrations as possible to avoid collecting large datasets of destructive collision examples. We explore whole-body proximity sensors as inductive biases for learning reflexive collision-dodging behavior and find that lossy sensors with low resolution were stronger biases than high-resolution coverage. We also found that, perhaps unintuitively, anticipating when and where contact happens on the body of a robot reduced its learning sample efficiency. We investigated eight candidate representations in a  reinforcement learning environment, which included latent compression, future contact anticipation, lossy corruption, and privileged state estimation. The candidate representations were tested by training over 200 humanoid robots to play a simplified version of dodgeball end-to-end under an intentionally constrained sample and compute budget.

\end{abstract}

\section{Introduction}

\begin{wrapfigure}{r}{0.33\linewidth}
    \centering
    \vspace{-2em}
    \includegraphics[width=\linewidth]{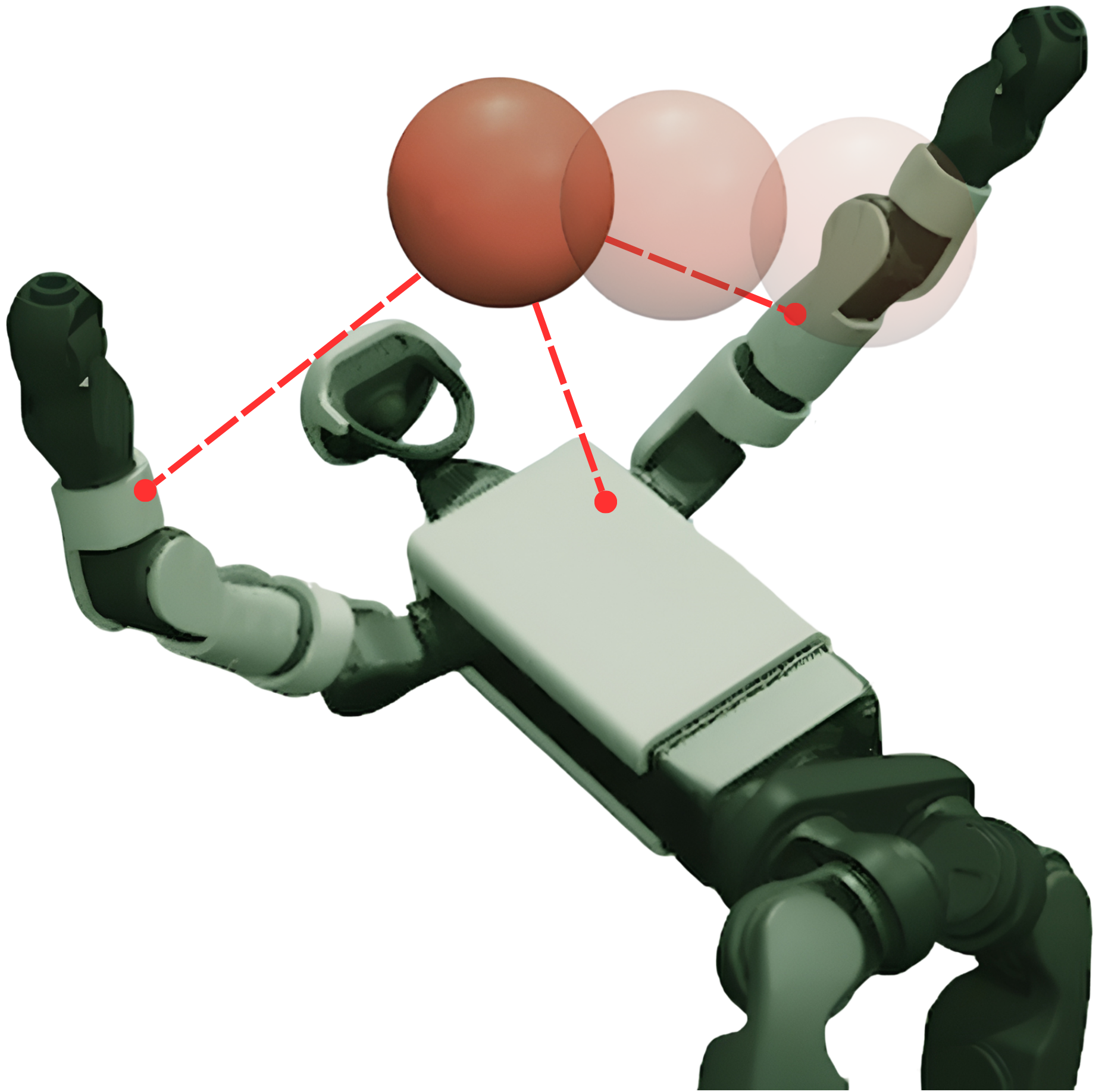}
    \caption{A humanoid robot dodges an incoming projectile using wide coverage, low resolution proximity sensors.}
    \label{fig:cover}
\end{wrapfigure}

Robots deployed in unstructured environments must avoid collisions that could damage themselves or others.
Data-driven models have enabled manipulation and locomotion tasks to be performed in unstructured environments \citep{kim2024openvla, intelligence2025pi_}, but learning collision avoidance through data is challenging because collision data can be damaging and costly to collect. This expensive data constraint alone motivates the need for sample-efficient ways to learn safe, reflexive behavior in as few demonstrations as possible.


Forecasting future camera observations using generative world models
emerged as a strategy for safely experiencing potentially dangerous scenarios \citep{agarwal2025cosmos, assran2025v}, and forecasting has also been shown to dramatically improve task performance in dynamic tasks such as sports \citep{syed2026intercepting}. However, occlusions and partial observability are a notorious problem for camera-based contact detection \citep{navarro2021proximity}, causing world models to rely on heuristic estimates of contact \citep{liu2025ego}. We therefore look to alternative sensors than cameras that are designed to combat occlusions.

Egocentric proximity sensors exist as solutions to occlusion and partial observability due to their ability to sense directly from the surface of a robot \citep{scholz2024sensor} and have been extensively used for safe collision avoidance \citep{navarro2021proximity}. A large variety of sensing methods power these whole-body proximity sensor constellations, such as light ranging, capacitive sensing, acoustic sensing, and more \citep{scholz2024sensor}. Despite the large variety of options, there is not a consensus on what properties of proximity sensors are ideal for sample-efficient learning, such as range or signal representation, and world models have yet to integrate such proximity sensors. Thus, we seek to answer the following questions for learning collision avoidance: How should a proximity sensor's signal be shaped, and does anticipating contact help?

The goal of this paper is two-fold: 1) to investigate how whole-body proximity sensor observations can be best represented as inductive biases for sample-efficient collision avoidance, and 2) to investigate if predicting future contact observations can lead to improved sample efficiency for learning collision avoidance.
The contributions of this work are as follows: 
\begin{enumerate}
    \item We propose a design heuristic for using proximity sensors as sample-efficient inductive biases, which favors wide, overlapping coverage of low-resolution sensors and low signal dimensionality. We further show how a high-resolution proximity sensor can mimic a low-dimensional sensor using a simple lossy signal minimization to improve sample efficiency.
    \item We propose that anticipating when and where contact will happen can create sample-efficiency deficits for learning collision avoidance, which we validate using a proximity world model (PWM) that forecasts future observations from whole-body proximity sensors to anticipate contact.
    \item We rigorously ablate sensor observation representation, including signal processing, forecasting, and hardware properties, in a simulated collision avoidance task with a Unitree humanoid H1-2 robot equipped with 64 proximity sensors.
\end{enumerate}

\section{Proximity Signal Representation}


Egocentric proximity sensors, also sometimes referred to as artificial skins, have been used for collision avoidance using time-of-flight sensors that pulse light or sound waves and measure distance by how long it takes the waves to return \citep{borelli2026sestosenso, kim2024armor, escobedo2021contact}, capacitive sensors that measure distance by how long it takes to charge a conductive material based on nearby objects \citep{kohlbrenner2026gentactprox, liu2023self, moon2021real}, or stereo cameras that measure depth from pixel disparity \citep{hu2025vlsa}. However, despite the wide variety of applicable proximity sensors, no metric exists to quantify how much of a proximity sensor’s signal is task-relevant for collision avoidance. We therefore start with a generic form for proximity sensors that is agnostic to an underlying sensing mechanism.


\subsection{Generic Proximity Sensor}

Consider a proximity sensor that activates when any point on a nearby object intersects with the proximity sensor’s coverage area. A proximity sensor’s coverage normally depends on its underlying mechanisms, but we assume our modality-agnostic proximity sensor is limited only by range and can have either full spherical coverage around the origin or a rectangular window within that sphere, defined by a horizontal and vertical field-of-view (FoV). Now consider the sensing coverage being sectioned into a grid of receptive fields that can each individually activate. The receptive fields set the resolution of the sensor, which we will denote $W$ as the number of receptive field regions along the horizontal FoV and $H$ as the number of receptive field regions along the vertical FoV.

Let a proximity sensor suite comprise $S$ sensors. The observation $x$ made by the sensors at time $t$ is thus $x_t\in \mathbb{R}^{H\times W\times S}$. We assume all sensors are ideal, meaning they have zero phase offset, latency, and noise, because we assume these are factors an engineer would choose to eliminate and are not design choices for sample efficiency. We will use this generic proximity sensor notation for the remainder of this work.

\begin{table}[t]
    \centering
    \renewcommand{\arraystretch}{1.25}
    \setlength{\tabcolsep}{6pt}

    \begin{tabular}{
        |p{0.2\linewidth}
        |p{0.3\linewidth}
        |c
        |c
        |c
        |
    }
        \hline
        \textbf{Representation $u_t$}
        & \textbf{Description}
        & \textbf{Distance}
        & \textbf{Direction}
        & \textbf{Dimension}\\
        \hline

        $x_t$
        & Raw sensor observation.
        & -- & --
        & $W \times H \times S$ \\
        \hline

        $f_\text{tp}(x_t)$
        & Localizes the nearest object within sensing coverage relative to the base of the robot.
        & $\uparrow$
        & $\uparrow$
        & $\downarrow\;(3)$ \\
        \hline

        $f_\text{min}(x_t)$
        & Returns only the closest range measurement per sensor.
        & --
        & $\downarrow$
        & $\downarrow\;(S)$\\
        \hline

        $f_\psi(x_t)$
        & Returns a latent observation.
        & $\approx$
        & $\approx$
        & $\downarrow$ \\
        \hline

        $f_\text{bin}(x_t)$
        & Returns only binary activations per signal.
        & $\downarrow$
        & --
        & -- \\
        \hline

        $\text{PWM}_0(x_t)$
        & Returns $[f_\text{tp}(x_t), f_{\text{tp}}(\hat{x}_{k}), c_{k}]$.
        & $\uparrow$
        & $\uparrow$
        & $\downarrow (7)$ \\
        \hline

        $\text{PWM}_{\text{min}}(x_t)$
        & Returns $[f_\text{min}(x_t), f_{\text{min}}(\hat{x}_{k}), c_{k}]$.
        & --
        & $\downarrow$ 
        & $\downarrow(2S+1)$\\
        \hline

        $\text{PWM}_\psi(x_t)$
        & Returns $[f_\psi(x_t), f_\psi(\hat{x}_{k}), c_{k}]$.
        & $\approx$
        & $\approx$
        & $\downarrow$ \\
        \hline

    \end{tabular}

    \caption{
        Investigated representations of sensor observation $x$ that modify the robot's access to location-revealing information.
        We assume the encoder representation $f_\psi$ performs compression and is well trained,
        such that $W' \times H' \times S' \times e < W \times H \times S$, while retaining relevant information about nearby object distance and direction.
    }
    \label{tab:representations}
\end{table}

\subsection{Sample-Constrained Usable Information}
\label{sec:usable_information}

The sample efficiency of collision avoidance learning can be interpreted through usable information theory \citep{xu2020theory}. Our motivating question is:
\emph{under a capacity-constrained learner and a limited training
budget, which representation of proximity sensor data makes collision-relevant information easiest to exploit?}

Let $X$ denote the random variable corresponding to a raw proximity
observation in the sample space $\mathcal{X}$, let $Y$ denote a collision-relevant target variable in $\mathcal{Y}$, and let $\mathcal{V}$ be a
restricted family of predictors of $Y$ used by the robot. $\mathcal{V}$ is a restricted family because the agent must predict $Y$ in finite time (i.e., the robot cannot take its time to compute how to dodge an impending collision). The predictive
$\mathcal{V}$-information $I_\mathcal{V}$ from $X$ to $Y$ is
\begin{equation}
I_{\mathcal{V}}(X \rightarrow Y)
=
H_{\mathcal{V}}(Y \mid \varnothing)
-
H_{\mathcal{V}}(Y \mid X),
\label{eq:vinfo}
\end{equation}
where $H_\mathcal{V}(Y\mid X)$ is the conditional $\mathcal{V}$-entropy,
\begin{equation}
H_{\mathcal{V}}(Y \mid X)
=
\inf_{q \in \mathcal{V}}
\mathbb{E}_{x,y\sim \mathcal{X}, \mathcal{Y}}
\left[-\log q[x](y)\right],
\end{equation}
and $H_{\mathcal{V}}(Y \mid \varnothing)$ is the $\mathcal{V}$-entropy when the predictor is given no observation of $X$. Applying a deterministic transformation $f:\mathcal{X}\rightarrow\mathcal{X}$ on $X$ would not increase the Shannon mutual information with $Y$, but under a restricted compute budget, a representation $f(X)$ that makes it easier for the robot to exploit $Y$ has a higher $\mathcal{V}$-information with $Y$ \citep{xu2020theory}. We therefore seek representations that satisfy
\begin{equation}
I_{\mathcal{V}}(f(X) \rightarrow Y)
>
I_{\mathcal{V}}(X \rightarrow Y)
\label{eq:vinfo_goal}
\end{equation}
for the learner class and collision avoidance task. Because an agent with greater exploitability on $Y$ may reach high reward states quicker, we treat
\cref{eq:vinfo_goal} as an interpretive lens and evaluate its implications empirically through downstream collision-avoidance learning. Optional ignorance is not a necessary assumption for this study; in fact, predictors that perform \textit{worse} with additional information may indicate harmful information for learning is presented.

This predictive $\mathcal{V}$-information perspective motivates the representations laid out in \cref{tab:representations}. The sensor-level transformations $f_{\min}$ and $f_{\mathrm{bin}}$ deliberately discard aspects of $x_t$, $f_{\psi}$ transforms $x_t$ into a lower-dimensional latent representation intended to preserve collision-relevant structure, and the PWM forecasts future contact and future observations from $x_t$ at time $k$. If contact is predicted to happen, $k$ is the time of contact, otherwise it is a fixed time $R$ ahead of the current time. We also consider an oracle that reveals privileged information about the incoming object’s state as an upper-bound representation. Together, these representations provide insight into how best to transform raw sensory data for sample efficiency and what information is most relevant to preserve.

For a given representation $U$ and corresponding task success curve $S_U(n)$, we use steps to a given success rate $N_\alpha(U)$ with $\alpha$ steps and normalized area under the curve (NAUC),
\begin{equation}
\text{NAUC}(U)=\dfrac{1}{N}\int_{0}^NS_U(n)dn,
\end{equation}
as metrics to evaluate which representations are more readily exploitable by $\mathcal{V}$.

\section{Contact Anticipation}

\begin{figure}
    \centering
    \includegraphics[width=\linewidth]{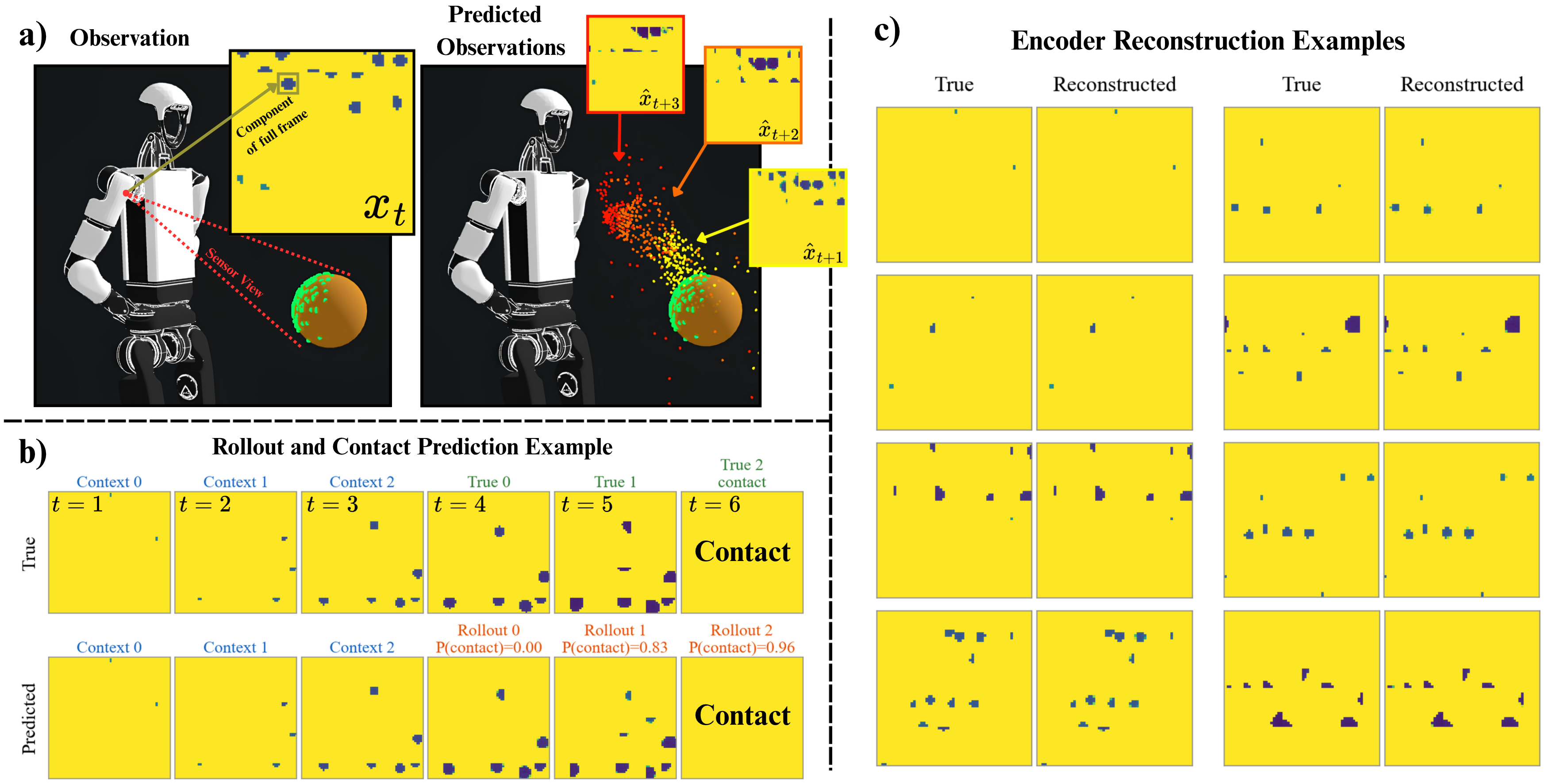}
    \caption{a) The PWM sensor observations are shown here as side-by-side frames of individual sensor depth images, where each sensor occupies an $W\times H$ grid of pixels in the observation frame. The green point cloud hits are points on the ball that cross a sensor's receptive field. When we roll out future observations, we can also visualize them in world space as a point cloud. Here we see the next three predicted observations $\hat{x}_{t+1:t+3}$. b) shows how a rollout will end when contact is predicted, and c) demonstrates the accuracy of the encoder and decoder in $\text{PWM}_\psi$ from random samples of a validation set. }
    \label{fig:examples}
\end{figure}

Let \(c_t\in\{0,1\}\) denote the contact state at time \(t\). The discrete objective of contact anticipation is to predict the conditional probability distribution of future contact at time $k$ based on previous observations $o_{\leq t}$.
\begin{equation}
    p(c_{t+1:t+k}\mid o_{\leq t}).
\end{equation}
Early approaches estimated the velocity of nearby objects using camera and tactile sensors to predict time-to-contact \citep{roncone2016peripersonal}, but this approach assumes dynamics of the nearby object. \citet{he2025foar, liu2025ego} proposed instead using a learned contact prediction model $p_\theta$ conditioned on camera images as prior observations to predict future contact states. \citet{zhou2025chronodreamer, team2026n_0, qin2026dextouch} further conditioned $p_\theta$ on tactile sensor observations and predicted future actions $\hat{a}_{t+1:t+k}$
\begin{equation}
    \hat{c}_{t+1:t+k} \sim p_\theta(c_{t+1:t+k}\mid o_{\leq t}, \hat{a}_{t+1:t+k}),
    \label{eq:contact_anticipator}
\end{equation}
However, their approach was validated in a manipulation setting, and anticipating future contact states at a whole-body scale for collision avoidance using proximity sensors has not been explored.

\subsection{Proximity World Model}

\begin{wrapfigure}{r}{0.3\linewidth}
    \centering
    \includegraphics[width=\linewidth]{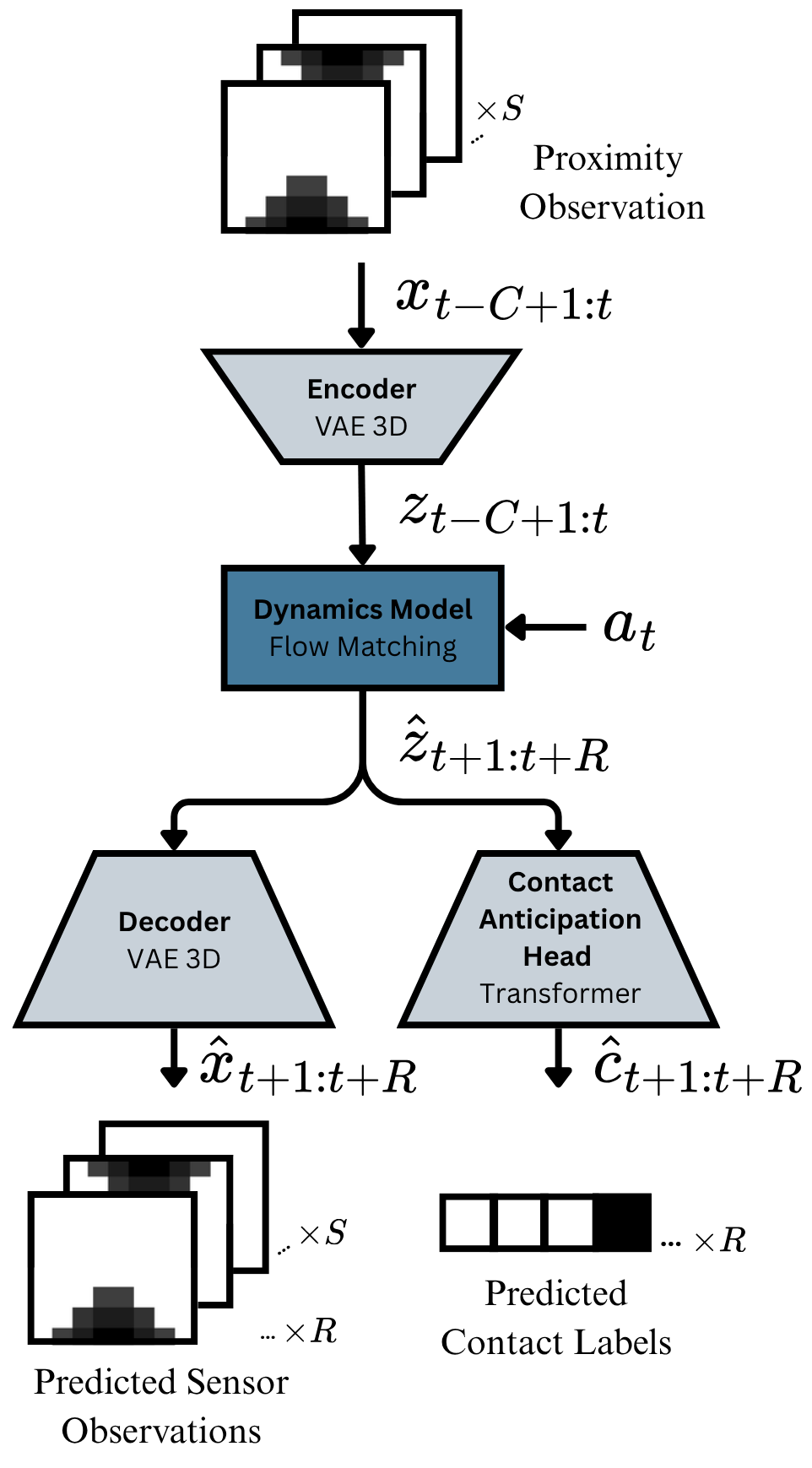}
    \caption{PWM architecture.}
    \label{fig:flow_diagram}
\end{wrapfigure}

We propose a proximity world model (PWM) that first generates future latent sensor observations, then uses those observations to predict contact. Following previous generative world model work \citep{agarwal2025cosmos}, our world model comprises an encoder, dynamics model, and decoder to generate future observations, as shown in \cref{fig:flow_diagram}. We introduce an additional contact prediction module to anticipate which predicted future observations will lead to contact.

Our predictive world model starts by mapping $x_t$ into a smaller dimensional latent vector $z_t\in \mathbb{R}^{H'\times W' \times S' \times e}$ through the encoder 
\begin{equation}
    \tilde{z}_t\sim    q_\psi(z_t\mid x_t)
\end{equation}
to improve sample efficiency during training, where $q_\psi$ is a learned posterior distribution and $e$ is an embedding dimension. At inference time, the encoder deterministically returns the latent observation as $z_t:=f_\psi(x_t)$. The decoder model conversely samples a likely sensor observation conditioned on a given latent observation. 

Next, the dynamics model \(r_\phi\) samples a possible sequence of future latent sensor observations induced by the robot's latest action
\begin{equation}
\hat{z}_{t+1:t+k}
\sim
r_\phi(
z_{t+1:t+k}
\mid
z_{\le t},a
).
\end{equation}

Accurate prediction of these future observations requires the world model to capture the relative dynamics between the robot and its environment. We then modify \cref{eq:contact_anticipator} to get the conditional probability distribution of contact conditioned on the predicted future observations and sample from a learned contact anticipation model $p_\eta$ to get a candidate set of future contact states:
\begin{equation}
\hat{c}_{t+1:t+k}
\sim
p_\eta(
c_{t+1:t+k}
\mid
\hat{z}_{t+1:t+k}, z_{\le t},a
).
\end{equation}

Thus, sampling a future trajectory from the world model and subsequently predicting its associated contact states provides a sample-based approximation to contact anticipation under a candidate action sequence. \cref{fig:examples} shows examples of how the current sensor observation is constructed using the ray sensors, as well as how future observations are propagated through the world model during a roll out. The PWM used in this work was pretrained on a sample dataset of 6000 trajectories collected via expert demonstration; see \cref{app:wm} for implementation details. For the same reason that sensors are represented without noise or latency,  the large training data is intended to make the PWM well adjusted as an \textit{idealized} contact predictor, that, while not perfect, serves as a higher-bound indicator of the influence of contact anticipation for sample-efficiency than a highly noisy PWM. It is assumed that contact prediction inaccuracy is ideally minimized in deployment. 
\section{Experiments}
\begin{figure}
    \centering
    \includegraphics[width=\linewidth]{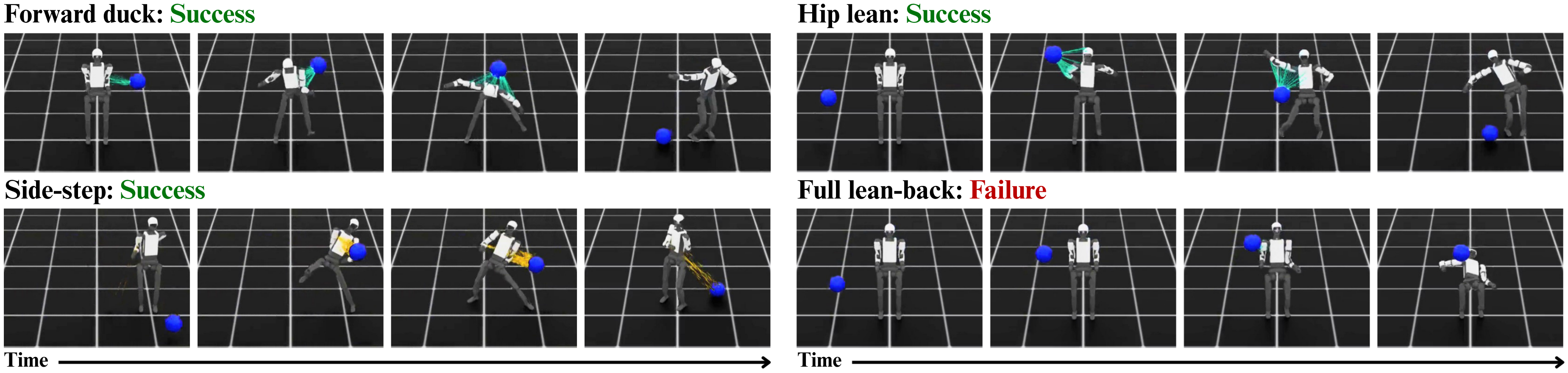}
    \caption{The robot is rewarded for learning collision-avoidant behavior and maintaining posture.}
    \label{fig:sequence}
\end{figure}

\begin{wrapfigure}{r}{0.46\linewidth}
    \centering
    \includegraphics[width=\linewidth]{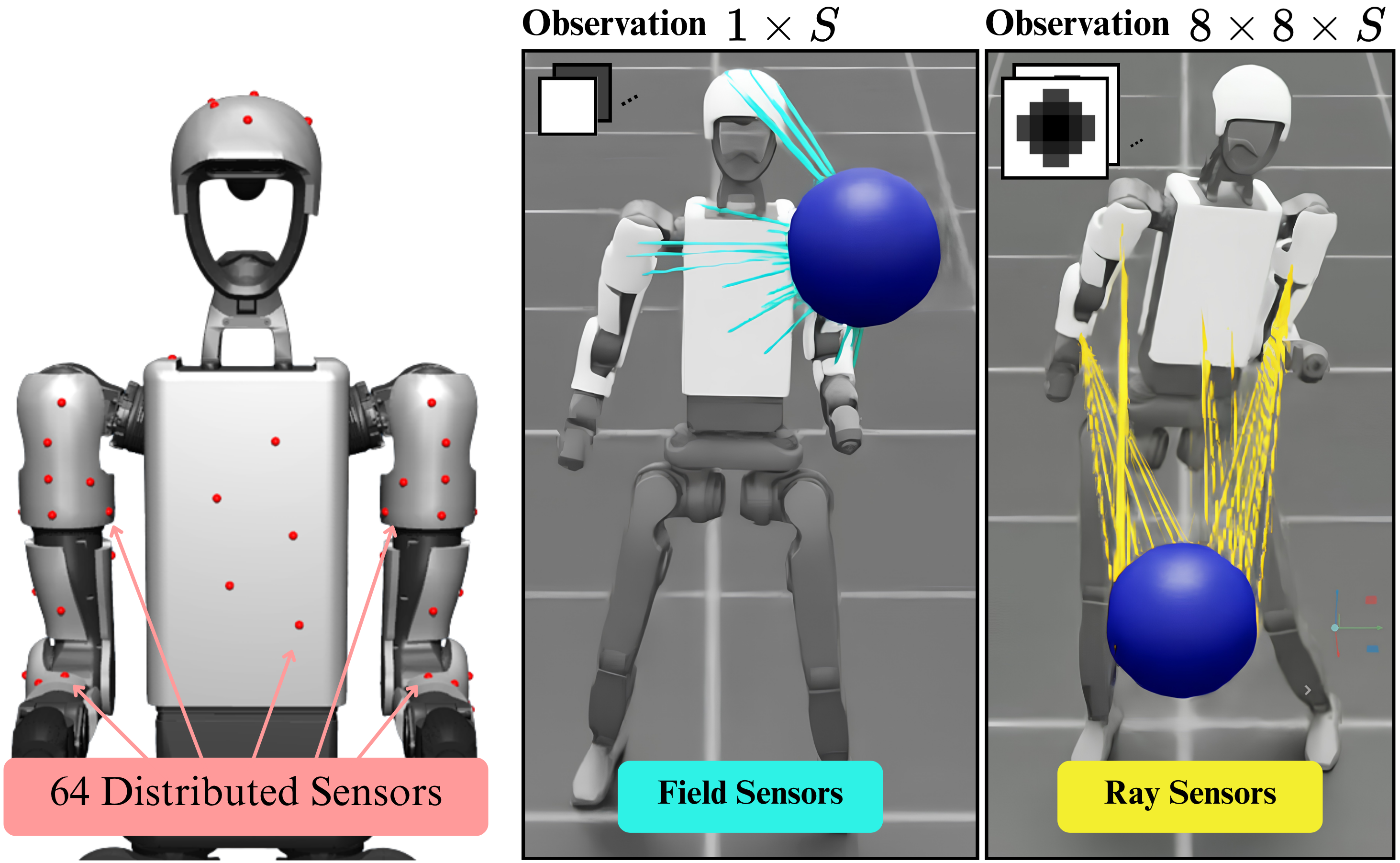}
    \caption{Example of distributed field-sensor activations (left) and ray-sensor activations (right). Lines connect to detected objects.}
    \label{fig:robot}
\end{wrapfigure}

We investigated the signal representations in \cref{tab:representations} empirically in a simulated collision avoidance task using a Unitree H1-2 humanoid robot equipped with 64 proximity sensors that were randomly distributed using GenTact \citep{kohlbrenner2024gentact} to supply ample coverage of the surrounding space of the robot, as shown in \cref{fig:robot}. The humanoid robot was trained using proximal policy optimization (PPO) \citep{schulman2017proximal} and was tasked with avoiding collisions with independently thrown balls launched at random times 1-2 seconds apart and set on a collision course with the robot. The balls had a constant radius of $\qty{15}{cm}$, spawned between 2-3 m away from the robot, between 1-3 m above ground level, and had an initial speed between 4-6 m/s. \cref{fig:sequence} shows a few successful and unsuccessful collision-avoidance policies for this task.

\paragraph{Task:} The policy $\pi_\theta(a_t|o_t)$ outputs target joint positions $q^*_t \in \mathbb{R}^{21}$, which are converted into motor torques $\tau_t$ via a low-level PD controller: 
\begin{equation}
\tau_t = K_p (q^*_t - q_t) + K_d (\dot{q}^*_t - \dot{q}_t),
\end{equation}
where $K_p$ and $K_d$ are the proportional and derivative gains, and joint velocities $\dot{q}^*_t$ are approximated by zeros or finite differences.
The policy commands to maximize the expected cumulative reward to optimize towards the optimal parameters $\theta^*$:
\begin{equation}
 J(\theta)=\mathbb{E}_{r\sim p_{\pi_\theta}}\left[\sum_{t=0}^{T} \gamma^t r_t\right]
\end{equation}
where $\gamma$ is the discount factor and $r_t$ is the reward function.

\paragraph{Termination:} An episode can end under four conditions: 1) the ball collides with the robot, 2) the robot's root point (at the pelvis) falls below a 40 cm threshold from the ground, 3) the robot's root point pose exceeds an $80^\circ$ angle from upright, and 4) the robot successfully survives for three seconds.

\paragraph{Rewards:}

The policy is optimized at \(60\,\mathrm{Hz}\) (\(\Delta t=1/120\,\mathrm{s}\) with decimation \(2\)). Each step returns
\begin{equation}
r_t
= w_h\exp\!\big(-\alpha(h_t-h^\star)^2\big)
+ w_a
- w_d\|\Delta p_t\|_2^2
- w_e\|\tau_t\|_1.
\label{eq:reward}
\end{equation}
Here \(h_t\) is the root height in meters, \(h^\star\) is a target standing height, \(\Delta p_t\in\mathbb{R}^2\) is the horizontal displacement from the reset pose, and \(\tau_t\) is the vector of applied joint torques. The Gaussian rewards an upright posture, and \(w_a\) is received on every step until termination, so longer episodes are the objective that produces dodging. The drift and torque terms suppress leaving the spawn point and high-frequency motion. \cref{eq:reward} depends only on the root state and the joint torques, so the reward does not favor a particular proximity-sensor representation. The following hyperparameters were found to be sufficient for learning: \(w_h=10\), \(\alpha=5\), \(h^\star=1.04\,\mathrm{m}\), \(w_a=2\), \(w_d=1\), and \(w_e=0.01\).

\paragraph{Observations:} In addition to the sensor observation, the observation space is comprised of base angular velocity $ \omega_b$, relative joint positions $q_r$, joint velocities $\dot{q}$, gravity vector projected in the base frame $g_b$, and the previous action for one-step action memory $a_{t-1}$, as
\begin{equation}
    o_t=[x, \omega_b, g_b, q_r, \dot{q}, a_{t-1}].
\end{equation}

Additionally, we adopt an asymmetric actor-critic structure \citep{pinto2017asymmetricactorcriticimagebased} to improve training stability, where the critic receives a privileged linear base velocity observation $v_b$. The actor and critic are each two-layer feedforward networks with 32 units per layer. We chose this fixed, low-capacity architecture so differences in task performance can be attributed to the proximity sensor observation rather than model expressivity \citep{bengio2013representation}.

\paragraph{Steps:} We report sample efficiency in PPO learning iterations. Each iteration collects 16 steps from 4,096 parallel environments (65,536 environment interactions, reused for 5 epochs in minibatches of 16,384) at a control rate of 60 Hz. All training runs share this budget, so an $m\times$ reduction in iterations is an $m\times$ reduction in environment interactions.

\begin{figure}[t]
    \centering

    \begin{subfigure}{\linewidth}
        \centering
        \includegraphics[width=\linewidth]{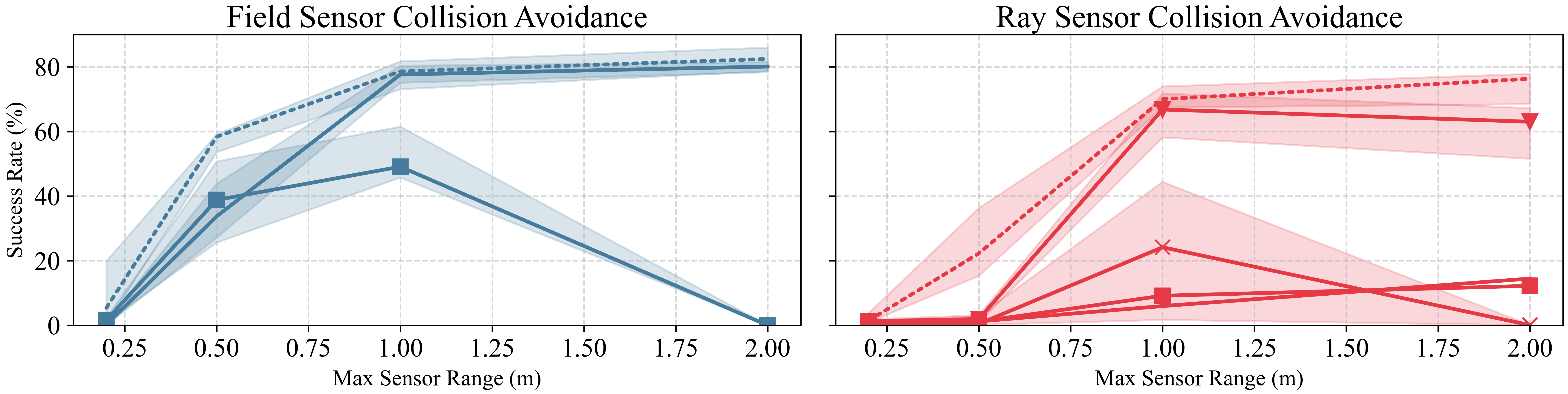}
        \phantomcaption
        \label{fig:range_ablation_results-a}
    \end{subfigure}


    \begin{subfigure}{\linewidth}
        \centering
        \includegraphics[width=\linewidth]{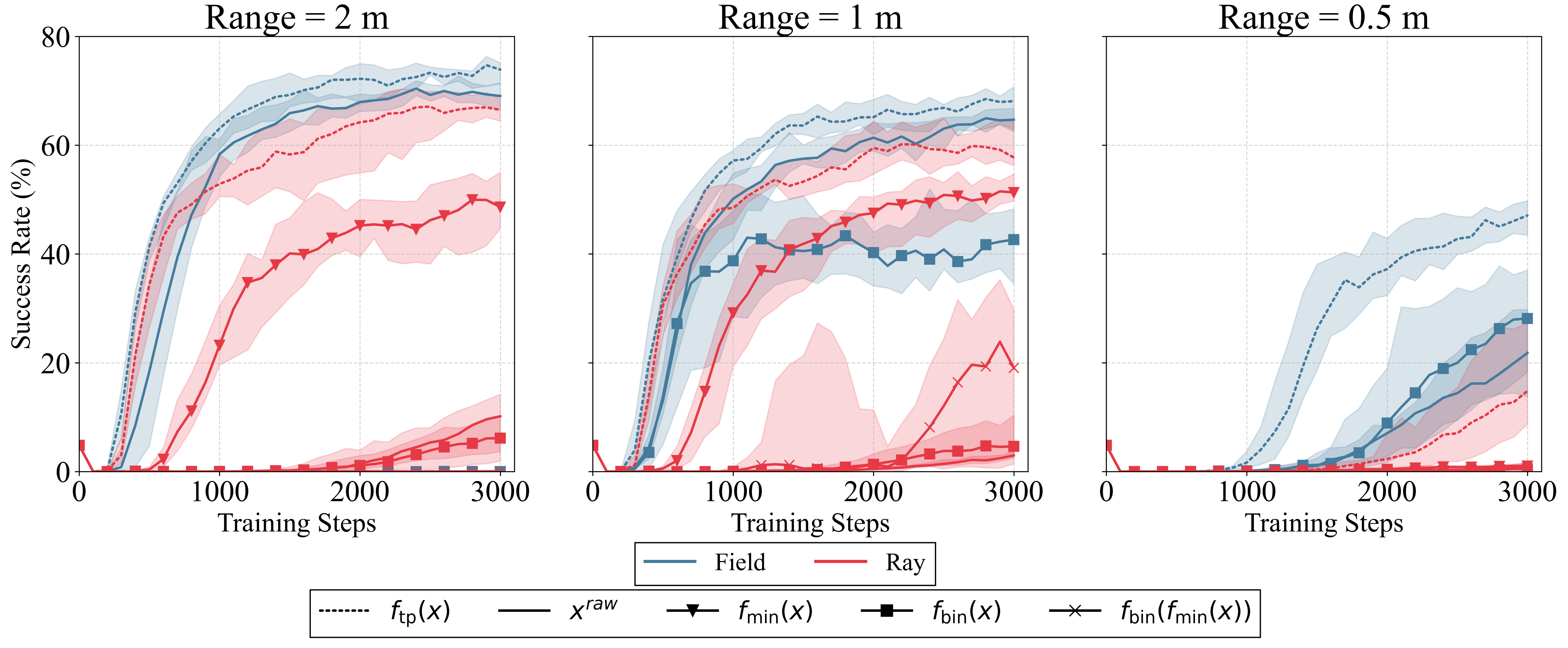}
        \phantomcaption
        \label{fig:step_ablation_results-b}
    \end{subfigure}

    \caption{
        Sensor representation ablation results.
        Top: total task success for field and ray sensors across ranges of
        0.2, 0.5, 1, and 2 meters.
        Bottom: task success over training steps for the field and ray sensor
        representations.
    }
    \label{fig:sensor_ablation_results}
\end{figure}
\subsection{Sensor Representation Ablation}
\label{sec:sensor_rep}

Inspired by the current dichotomy of the proximity sensing field, we define two extremes of the generic proximity sensor to use as test subjects: \textbf{field sensors} (low-resolution), which we define as a single receptive field (\(H, W=1\)) with wide coverage and no explicit indication of where an object lies within the field of view, and \textbf{ray sensors} (high resolution), which use an \(8\times8\) sensing grid over a narrow \(45^\circ\) field of view, where each signal in the grid is associated with a unique heading direction. We perform 10 seeded end-to-end training runs of the dodgeball task with a maximum compute budget of 3000 steps per run. \cref{fig:range_ablation_results-a} reports the median and interquartile range (IQR) of task success across seeds for these sensors and the signal representations in \cref{tab:representations}, and \cref{fig:step_ablation_results-b} reports the sample efficiency of each representation in task success over training steps.

\textit{Perfect Localization Oracle $f_{tp}$:} We found that robots that could observe the true relative position of the incoming ball $f_{tp}$ served as an upper-bound representation for both the field sensor and the ray sensor, achieving a NAUC of 0.584 at 2 m range. $f_{tp}$ is believed to be the strongest representation because it compactly encodes both distance and direction in a three-dimensional signal.

\textit{Raw Distance Observation $x_t$:} Robots with raw \textit{field} sensor observations achieved a NAUC of 0.522 at 2 m range, which is $89.3\%$ as strong as those with $f_{tp}(x_t^{\text{field}})$ across all sensing ranges, requiring no modification to their representation for their performance. Although each field sensor provides no explicit directional information, we believe its distributed activations allowed the policy to infer object direction from the sensor constellation itself. In contrast, policies trained on raw ray sensor observations only reached a maximum NAUC of 0.02.

\textit{Minimum Distance Observation $f_{\text{min}}$: } Applying the minimum-distance transformation \(f_{\min}\) substantially improved both final performance and sample efficiency of the robots with ray sensors, despite discarding the within-sensor direction of the detected object. At $\qty{1}{m}$ sensing range, $f_\text{min}(x^{\text{ray}})$ outperformed raw ray sensor observation NAUC by 5400\%. By retaining only the nearest range measurement from each sensor, \(f_{\min}\) effectively converted each ray sensor into a field-like observation within its sensing coverage. Additionally, $f_{\text{min}}$ provides insight into how coverage itself influenced the robot's ability to learn. Robots with raw field sensors achieved an average NAUC 0.138 points higher than those with $f_\text{min}(x^{\text{ray}})$ observations, despite $f_\text{min}(x^{\text{ray}})$ observations occupying the same range and approximate signal properties to raw field sensors. Similarly, robots that observed $f_\text{tp}(x_t^\text{field})$ performed an average NAUC of 0.12 higher than those with $f_\text{tp}(x_t^\text{ray})$. We attribute these marginal gains to the increased coverage of the field sensors, which makes these two examples the only comparable representations between sensor types. Field sensors were not evaluated with \(f_{\min}\) due to only containing a single signal by default.

\textit{Binary Detection Observation $f_{\text{bin}}$: } We observed that at the $\qty{2}{m}$ sensing range, the $f_{\text{bin}}$ representation would saturate well before contact for both sensor types. However, $f_{\text{bin}}(x^\text{field})$ outperformed direct field proximity measurements at the $\qty{0.5}{m}$ sensing range, suggesting that distance becomes less important in the moments before contact.

\textit{Latent Observation $f_\psi(x)$:} Learned compression with \(f_\psi\) improved the sample efficiency relative to raw ray observations, but did not perform as well as \(f_{\min}\). This can be understood by the dimensionality of $f_\psi(x)$ being $f_\psi(x)$ being $\dim{x}=4096>\dim{f\psi(x)}=512>\dim{f_{\min}(x)}=64$, where the encoder was pretrained to output embeddings in the dimension of $H', W'=2$, $S'=16$, and $e=8$, resulting in an embedding that is one-eighth the size of the raw observation.

\subsection{Contact Prediction Ablation}

\begin{wrapfigure}{r}{0.44\linewidth}
    \centering
    \includegraphics[width=\linewidth]{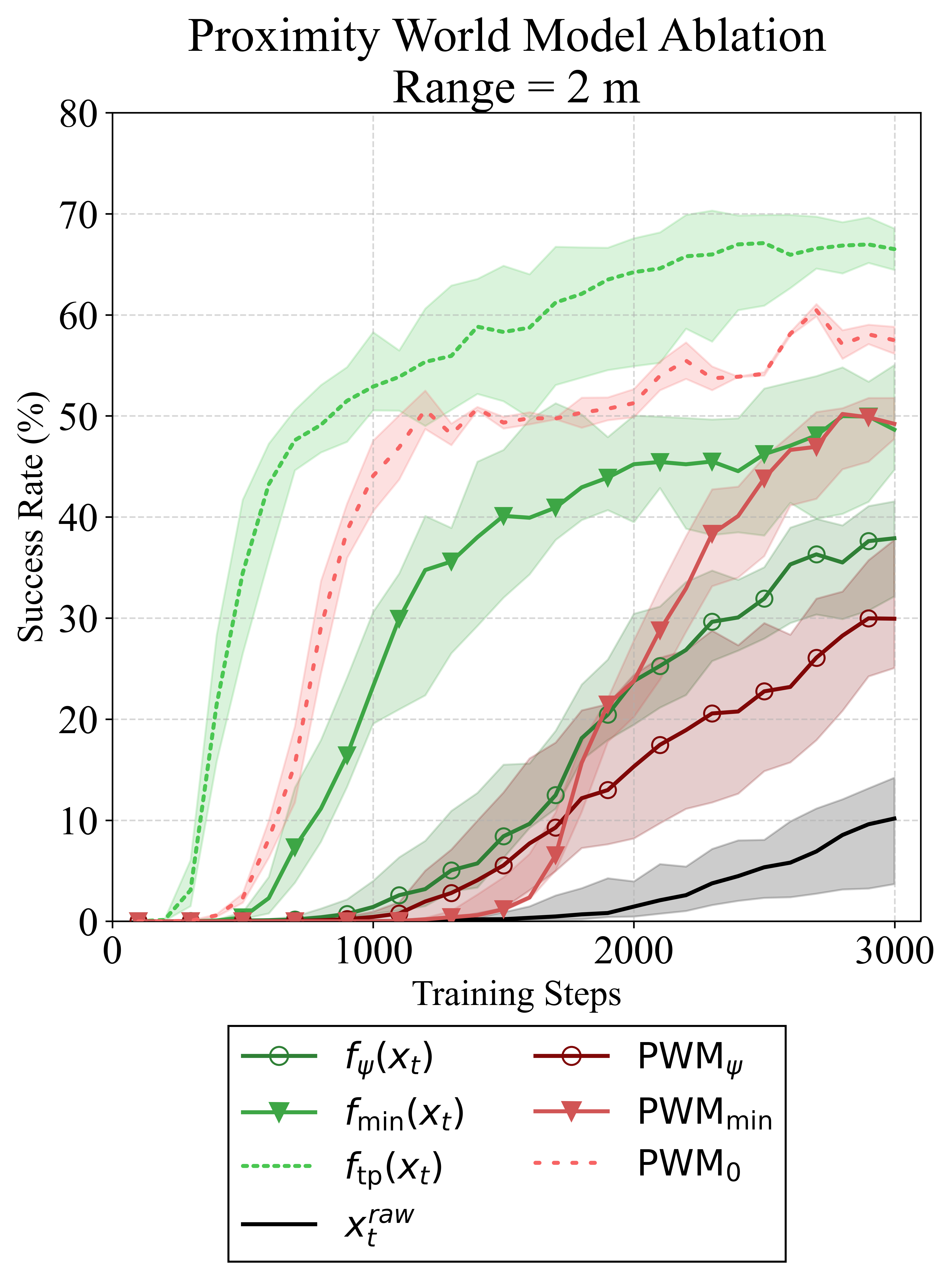}
    \caption{Robots that could anticipate contact (red) consistently underperformed variants that used only current observations (green).}
    \label{fig:wm_results}
\end{wrapfigure}

\begin{table}[t]
    \centering
    \resizebox{\linewidth}{!}{
    \begin{tabular}{lccccccc}
        \toprule
        \multirow{2}{*}{\textbf{Sensor Observation}} &
        \multicolumn{3}{c}{\textbf{10\% Success}} &
        \multicolumn{3}{c}{\textbf{25\% Success}} &
        \multirow{2}{*}{\textbf{NAUC}} \\
        \cmidrule(lr){2-4}
        \cmidrule(lr){5-7}
        &
        \textbf{Steps} &
        \textbf{Speedup} &
        \textbf{w/ PWM} &
        \textbf{Steps} &
        \textbf{Speedup} &
        \textbf{w/ PWM} & \\
        \midrule

        $f_{\mathrm{tp}}(x_t)$
        & 338
        & \textcolor{green!50!black}{$8.8\times$ faster}
        & --
        & 427
        & \textcolor{green!50!black}{$>7.0\times$ faster}
        & --
        & 0.508 \\

        $\text{PWM}_0(x_t)$
        & 625
        & \textcolor{green!50!black}{$4.7\times$ faster}
        & \textcolor{red!50!black}{$1.9\times$ slower}
        & 770
        & \textcolor{green!50!black}{$>3.9\times$ faster}
        & \textcolor{red!50!black}{$1.8\times$ slower}
        & 0.39 \\

        $f_{\min}(x_t)$
        & 770
        & \textcolor{green!50!black}{$3.9\times$ faster}
        & --
        & 1026
        & \textcolor{green!50!black}{$>2.9\times$ faster}
        & --
        & 0.299 \\

        $\text{PWM}_\text{min}(x_t)$
        & 1738
        & \textcolor{green!50!black}{$1.7\times$ faster}
        & \textcolor{red!50!black}{$2.3\times$ slower}
        & 2024
        & \textcolor{green!50!black}{$>1.5\times$ faster}
        & \textcolor{red!50!black}{$2.0\times$ slower}
        & 0.158 \\

        $f_{\psi}(x_t)$
        & 1613
        & \textcolor{green!50!black}{$1.8\times$ faster}
        &--
        & 2082
        & \textcolor{green!50!black}{$>1.4\times$ faster}
        &--
        & 0.140 \\

        $\text{PWM}_\psi(x_t)$
        & 1725
        & \textcolor{green!50!black}{$1.7\times$ faster}
        & \textcolor{red!50!black}{$1.1\times$ slower}
        & 2663
        & \textcolor{green!50!black}{$>1.1\times$ faster}
        & \textcolor{red!50!black}{$1.3\times$ slower}
        & 0.099 \\

        $x_t$
        & 2970
        & baseline
        &--
        & $>3000$
        & baseline
        &--
        & 0.020 \\

        \bottomrule
    \end{tabular}
    }
    \caption{Median performance across sensor observation representations. Speedup represents how many less steps were needed to achieve success thresholds of 10\% and 25\% relative to raw ray sensor observations as the baseline. Observations with the PWM consistently reduced sample efficiency.}
    \label{tab:sample_efficiency}
\end{table}

We investigated using the PWM to improve the ray sensor's sample efficiency within the 3000-step allotted compute budget at the two-meter range, due to the large gap between $f_\text{tp}(x^{\text{ray}})$ and $x^\text{ray}$ at this range. We simulated the same dodgeball task, this time with five seeded training runs due to the increase compute cost for the PWM, where the robot could now additionally use the PWM to observe a future $k$ timestep. The PWM success rate during training time of this study is shown in red in \cref{fig:wm_results} and compared against the previous ray sensor representations from \cref{sec:sensor_rep}.

\textit{Future Insight Oracle $\text{PWM}_0$: } The oracle robot observed the true relative position of the ball in its current state and where it would be upon collision, but performed worse than the robot that only saw the current true position. Additionally, the oracle was unable to achieve any task success without additional information regarding where the ball came from.
This result suggests that providing information about the future collision state may not carry readily exploitable information for collision avoidance in this setting.

\textit{Current and Future Latent Observation $\text{PWM}_\psi$ and $\text{PWM}_\text{min}$:} Similarly to the oracle, providing predictive information about the future collision state harmed the agent's ability to learn collision avoidance quickly. Both the agent with $\text{PWM}_\psi$ and $\text{PWM}_\text{min}$ reached similar levels of task success as their non-predictive counterparts, but the additional information available reduced their ability to exploit task-relevant information in as many steps.

\section{Discussion}
Our ablations point to two consistent phenomena: \textbf{signal dimensionality} was the biggest predictor of sample efficiency, and \textbf{anticipating contact distracted} the robot from exploiting task-relevant information in the demonstrated close-by, single impending collision task. Despite the PWM being trained on thousands of expert demonstrations and having privileged pretraining data, it still did not beat a simple per-sensor minimum distance transformation.

Our findings are further supported by the neurosciences, where fast multisensory neurons encode a peripersonal space (PPS): a compact representation of body-centric receptive fields that activate based on the proximity of nearby objects \citep{bufacchi2018action}. Our field-sensor representation has a similar computational structure: individual sensors provide little spatial detail, yet their overlapping, body-distributed activations encode sufficient direction and distance for avoidance. One interpretation of our results is therefore that sample-efficient collision avoidance benefits less from reconstructing the external world than from constructing a compact, action-oriented representation of peripersonal threat. In our experiment, range was a fixed value throughout training, but it is well known that the PPS adapts in size based on the speed of nearby objects \citep{noel2018neural}. Future work should therefore explore a dynamically adapting \textit{range} for improving sample efficiency.

\section{Related Works}

\citet{zhang2025viability, armleder2025real, escobedo2021contact} used quadratic programming (QP) with predefined constraints on $o_t$ to achieve reactive control for collision avoidance. However, QP algorithms rely on careful tuning of their dynamics and object estimations to maintain safety. Instead of manually tuning a controller, \citet{yang2025deep, kim2024armor, wang2024duel} trained a policy with learned parameters $\theta$ to imitate expert demonstrations. Expert demonstrations can be expensive, so \citet{yang2026pac, xue2026collision} used reinforcement learning (RL) to train a reactive policy directly on observations guided by rewards, but this approach requires a simulation environment that needs to accurately emulate the real world for deployment and can require privileged information and an intractable number of trial-and-error training steps to be successful end-to-end on real data. Therefore, we believe that the goal is to reduce the number of training steps needed for effective collision avoidance to support a future where robots adapt to collisions end-to-end, in real time, and without relying on manual tuning, expert demonstrations, or privileged simulators.

\section{Conclusion}
We investigated eight candidate representations of proximity sensor signals for sample-efficient collision avoidance, including lossy transformations, latent compression, and world-model forecasting. We find in our experiments that wide coverage of low-resolution proximity sensors consistently resulted in higher sample efficiency against high-resolution alternatives, which are consistent with  $\mathcal{V}$-information theory as having higher $\mathcal{V}$-predictive relevant information for reflexive contact avoidance under sample and compute constraints. Additionally, providing robots with additional insight into when and where a collision would happen in the future resulted in consistently lower sample efficiency and did not overcome the cost of extra signal dimensionality. Therefore, our  findings suggest that for learning sample-efficient collision avoidance with egocentric proximity sensors: 1) wide-overlap, low-resolution sensing that preserves relative distance information and 2) conditioning learned policies only on current observations may be desirable for computationally constrained reflexive motion. 

\subsection*{AI use statement}

In this work, we used generative AI tools for assistance and feedback setting up our synthetic dataset collection, pipeline implementation between the PWM and training environment, and mathematical verification on the synthesis of our results. We have not used generative AI tools for data analysis, hypothesis creation, or interpretations of results, and assistance with mathematical proofs and translation are not applicable to this work. Additionally, we used generative AI tools as a method of discovering relevant sources, figure editing, grammar and spelling refinement, and software editing. We have reviewed all AI-assisted work by inspecting our datasets for accuracy via custom tests. We take responsibility for the final content of this work, including text, claims or artifacts produced with the aid of generative AI.

\subsubsection*{Acknowledgments}
Preliminary results of the sensor representation ablation were presented by \citet{kohlbrenner2026egocentric}, but this work further investigates the use of contact anticipation as an additional observation. This work utilized the Alpine High-Performance Computing resource at the University of Colorado Boulder. Alpine is jointly funded by the University of Colorado Boulder, the University of Colorado Anschutz, Colorado State University, and the National Science Foundation (award 2201538).

\bibliography{iclr2027_conference}

@article{armleder2025real,
  title={Real-Time Control of a Humanoid Robot for Whole-Body Tactile Interaction},
  author={Armleder, Simon and Bergner, Florian and Guadarrama-Olvera, Julio Rogelio and Nakanishi, Jun and Cheng, Gordon},
  journal={Advanced Intelligent Systems},
  volume={7},
  number={12},
  pages={e202500149},
  year={2025},
  publisher={Wiley Online Library}
}

@article{bengio2013representation,
  title={Representation learning: A review and new perspectives},
  author={Bengio, Yoshua and Courville, Aaron and Vincent, Pascal},
  journal={IEEE transactions on pattern analysis and machine intelligence},
  volume={35},
  number={8},
  pages={1798--1828},
  year={2013},
  publisher={IEEE}
}

@misc{pinto2017asymmetricactorcriticimagebased,
      title={Asymmetric Actor Critic for Image-Based Robot Learning}, 
      author={Lerrel Pinto and Marcin Andrychowicz and Peter Welinder and Wojciech Zaremba and Pieter Abbeel},
      year={2017},
      eprint={1710.06542},
      archivePrefix={arXiv},
      primaryClass={cs.RO},
      url={https://arxiv.org/abs/1710.06542}, 
}

@article{xue2026collision,
  title={Collision-Free Humanoid Traversal in Cluttered Indoor Scenes},
  author={Xue, Han and Liang, Sikai and Zhang, Zhikai and Zeng, Zicheng and Liu, Yun and Lian, Yunrui and Wang, Jilong and Liu, Qingtao and Shi, Xuesong and Yi, Li},
  journal={arXiv preprint arXiv:2601.16035},
  year={2026}
}

@inproceedings{kohlbrenner2024gentact,
  title={{GenTact Toolbox}: A Computational Design Pipeline to Procedurally Generate Context-Driven 3D Printed Whole-Body Artificial Skins},
  author={Kohlbrenner, Carson and Escobedo, Caleb and Bae, S Sandra and Dickhans, Alexander and Roncone, Alessandro},
  booktitle={2025 IEEE International Conference on Robotics and Automation (ICRA)},
  pages={4716--4722},
  year={2025},
  organization={IEEE}
}

@inproceedings{kohlbrenner2026gentactprox,
      title={Design, Mapping, and Contact Anticipation with 3D-printed Whole-Body Tactile and Proximity Sensors},
      author={Kohlbrenner, Carson and Soukhovei, Anna and Escobedo, Caleb and Nechyporenko, Nataliya and Roncone, Alessandro},
      booktitle={2026 IEEE International Conference on Robotics and Automation (ICRA)},
      year={2026}}

@article{liu2023self,
  title={A self-capacitance proximity e-skin with long-range sensibility for robotic arm environment perception},
  author={Liu, Zhe and Chen, Diansheng and Ma, Junlin and Jia, Daiyu and Wang, Tingyu and Su, Qile},
  journal={IEEE Sensors Journal},
  volume={23},
  number={13},
  pages={14854--14863},
  year={2023},
  publisher={IEEE}
}

@article{navarro2021proximity,
  title={Proximity perception in human-centered robotics: A survey on sensing systems and applications},
  author={Navarro, Stefan Escaida and M{\"u}hlbacher-Karrer, Stephan and Alagi, Hosam and Zangl, Hubert and Koyama, Keisuke and Hein, Bj{\"o}rn and Duriez, Christian and Smith, Joshua R},
  journal={IEEE Transactions on Robotics},
  year={2021},
  publisher={IEEE}
}

@inproceedings{escobedo2021contact,
  title={Contact anticipation for physical human--robot interaction with robotic manipulators using onboard proximity sensors},
  author={Escobedo, Caleb and Strong, Matthew and West, Mary and Aramburu, Ander and Roncone, Alessandro},
  booktitle={2021 IEEE/RSJ International Conference on Intelligent Robots and Systems (IROS)},
  pages={7255--7262},
  year={2021},
  organization={IEEE}
}

@article{roncone2016peripersonal,
  title={Peripersonal space and margin of safety around the body: learning visuo-tactile associations in a humanoid robot with artificial skin},
  author={Roncone, Alessandro and Hoffmann, Matej and Pattacini, Ugo and Fadiga, Luciano and Metta, Giorgio},
  journal={PloS one},
  volume={11},
  number={10},
  pages={e0163713},
  year={2016},
  publisher={Public Library of Science San Francisco, CA USA}
}

@article{kim2024armor,
  title={Armor: Egocentric perception for humanoid robot collision avoidance and motion planning},
  author={Kim, Daehwa and Srouji, Mario and Chen, Chen and Zhang, Jian},
  journal={arXiv preprint arXiv:2412.00396},
  year={2024}
}

@article{he2025foar,
  title={Foar: Force-aware reactive policy for contact-rich robotic manipulation},
  author={He, Zihao and Fang, Hongjie and Chen, Jingjing and Fang, Hao-Shu and Lu, Cewu},
  journal={IEEE Robotics and Automation Letters},
  volume={10},
  number={6},
  pages={5625--5632},
  year={2025},
  publisher={IEEE}
}

@article{kim2024openvla,
  title={{Openvla}: An open-source vision-language-action model},
  author={Kim, Moo Jin and Pertsch, Karl and Karamcheti, Siddharth and Xiao, Ted and Balakrishna, Ashwin and Nair, Suraj and Rafailov, Rafael and Foster, Ethan and Lam, Grace and Sanketi, Pannag and others},
  journal={arXiv preprint arXiv:2406.09246},
  year={2024}
}

@article{intelligence2025pi_,
  title={$\pi_{0.5}$: a Vision-Language-Action Model with Open-World Generalization},
  author={{Physical Intelligence} and Black, Kevin and Brown, Noah and Darpinian, James and Dhabalia, Karan and Driess, Danny and Esmail, Adnan and Equi, Michael and Finn, Chelsea and Fusai, Niccolo and others},
  journal={arXiv preprint arXiv:2504.16054},
  year={2025}
}

@article{moon2021real,
  title={Real-time obstacle avoidance using dual-type proximity sensor for safe human-robot interaction},
  author={Moon, Seung Jae and Kim, Jinsol and Yim, Hongsik and Kim, Yeeun and Choi, Hyouk Ryeol},
  journal={IEEE Robotics and Automation Letters},
  volume={6},
  number={4},
  pages={8021--8028},
  year={2021},
  publisher={IEEE}
}

@article{hu2025vlsa,
  title={{VLSA}: Vision-language-action models with plug-and-play safety constraint layer},
  author={Hu, Songqiao and Liu, Zeyi and Liu, Shuang and Cen, Jun and Meng, Zihan and Wang, Shihefeng and Li, Xiang and He, Xiao},
  journal={arXiv preprint arXiv:2512.11891},
  year={2025}
}

@article{lipman2022flow,
  title={Flow matching for generative modeling},
  author={Lipman, Yaron and Chen, Ricky TQ and Ben-Hamu, Heli and Nickel, Maximilian and Le, Matt},
  journal={arXiv preprint arXiv:2210.02747},
  year={2022}
}

@article{agarwal2025cosmos,
  title={Cosmos world foundation model platform for physical ai},
  author={Agarwal, Niket and Ali, Arslan and Bala, Maciej and Balaji, Yogesh and Barker, Erik and Cai, Tiffany and Chattopadhyay, Prithvijit and Chen, Yongxin and Cui, Yin and Ding, Yifan and others},
  journal={arXiv preprint arXiv:2501.03575},
  year={2025}
}

@article{assran2025v,
  title={{V-JEPA 2}: Self-supervised video models enable understanding, prediction and planning},
  author={Assran, Mido and Bardes, Adrien and Fan, David and Garrido, Quentin and Howes, Russell and Muckley, Matthew and Rizvi, Ammar and Roberts, Claire and Sinha, Koustuv and Zholus, Artem and others},
  journal={arXiv preprint arXiv:2506.09985},
  year={2025}
}

@article{zhou2025chronodreamer,
  title={Chronodreamer: Action-conditioned world model as an online simulator for robotic planning},
  author={Zhou, Zhenhao and Negrut, Dan},
  journal={arXiv preprint arXiv:2512.18619},
  year={2025}
}

@article{team2026n_0,
  title={{$N_0$-TWAM:} Scaling Tactile-Native World-Action Model for Contact-Rich Manipulation},
  author={Team, NeoteAI and Fudan TEAI Team and others},
  journal={arXiv preprint arXiv:2607.23783},
  year={2026}
}

@article{qin2026dextouch,
  title={DexTouch-WM: Learning Action-Conditioned Tactile World Models from Human Touch for Dexterous Robot Manipulation},
  author={Qin, Yan and Chen, Yue and Lin, Wenwei and Liu, Shujia and Lyu, Chuqiao and Su, Kailun and Yu, Chenze and Luo, Ping and Ding, Wenbo and Chen, Tianxing and others},
  journal={arXiv preprint arXiv:2609.20649},
  year={2026}
}

@article{kingma2013auto,
  title={Auto-encoding variational bayes},
  author={Kingma, Diederik P and Welling, Max},
  journal={arXiv preprint arXiv:1312.6114},
  year={2013}
}

@inproceedings{higgins2017beta,
  title={{$\beta$-VAE}: Learning basic visual concepts with a constrained variational framework},
  author={Higgins, Irina and Matthey, Loic and Pal, Arka and Burgess, Christopher and Glorot, Xavier and Botvinick, Matthew and Mohamed, Shakir and Lerchner, Alexander},
  booktitle={International conference on learning representations},
  year={2017}
}

@article{ghosh2019variational,
  title={From variational to deterministic autoencoders},
  author={Ghosh, Partha and Sajjadi, Mehdi SM and Vergari, Antonio and Black, Michael and Sch{\"o}lkopf, Bernhard},
  journal={arXiv preprint arXiv:1903.12436},
  year={2019}
}

@article{zhang2025viability,
  title={Viability-Preserving Passive Torque Control},
  author={Zhang, Zizhe and Wang, Yicong and Zhang, Zhiquan and Li, Tianyu and Figueroa, Nadia},
  journal={arXiv preprint arXiv:2510.03367},
  year={2025}
}

@inproceedings{
yang2025deep,
title={Deep Reactive Policy: Learning Reactive Manipulator Motion Planning for Dynamic Environments},
author={Jiahui Yang and Jason Jingzhou Liu and Yulong Li and Youssef Khaky and Kenneth Shaw and Deepak Pathak},
booktitle={2nd Workshop on Safe and Robust Robot Learning for Operation in the Real World},
year={2025},
url={https://openreview.net/forum?id=v2BNi6UQKr}
}

@article{wang2024duel,
  title={{DUEL}: Depth visUal Ego-motion Learning for autonomous robot obstacle avoidance},
  author={Wang, Naiyao and Zhang, Bo and Chi, Haixu and Wang, Hua and McLoone, Se{\'a}n and Liu, Hongbo},
  journal={The International Journal of Robotics Research},
  volume={43},
  number={3},
  pages={305--329},
  year={2024},
  publisher={SAGE Publications Sage UK: London, England}
}

@article{yang2026pac,
  title={{PAC-MAN}: Perception-Aware CBF-RL for Whole-Body Safety in Humanoid Dodgeball},
  author={Yang, Lizhi and Li, Junheng and Ames, Aaron D},
  journal={arXiv preprint arXiv:2607.28623},
  year={2026}
}

@article{schulman2017proximal,
  title={Proximal policy optimization algorithms},
  author={Schulman, John and Wolski, Filip and Dhariwal, Prafulla and Radford, Alec and Klimov, Oleg},
  journal={arXiv preprint arXiv:1707.06347},
  year={2017}
}

@article{liu2025ego,
  title={Ego-vision world model for humanoid contact planning},
  author={Liu, Hang and Gao, Yuman and Teng, Sangli and Chi, Yufeng and Shao, Yakun Sophia and Li, Zhongyu and Ghaffari, Maani and Sreenath, Koushil},
  journal={arXiv preprint arXiv:2510.11682},
  year={2025}
}

@article{scholz2024sensor,
  title={Sensor-enabled safety systems for human--robot collaboration: A review},
  author={Scholz, Constantin and Cao, Hoang-Long and Imrith, Emil and Roshandel, Nima and Firouzipouyaei, Hamed and Burkiewicz, Aleksander and Amighi, Milan and Menet, Sebastien and Sisavath, Dylan Warawout and Paolillo, Antonio and others},
  journal={IEEE Sensors Journal},
  volume={25},
  number={1},
  pages={65--88},
  year={2024},
  publisher={IEEE}
}

@inproceedings{borelli2026sestosenso,
  title={{SESTOSENSO}: Proxy-Tactile Perception for Intelligent and Safe Human--Robot Interaction},
  author={Borelli, Simone and Giovinazzo, Francesco and Grella, Francesco and Caroleo, Giammarco and Albini, Alessandro and Maiolino, Perla and Cannata, Giorgio},
  booktitle={2026 IEEE/ASME International Conference on Advanced Intelligent Mechatronics (AIM)},
  pages={1--8},
  year={2026},
  organization={IEEE}
}

@article{syed2026intercepting,
  title={Intercepting the future: Latent-space predictive world model for dynamic VLA manipulation},
  author={Syed, Shahram Najam and Jakobsson, Arthur and Hao, Haoran and Ichnowski, Jeffrey},
  journal={arXiv preprint arXiv:2606.02486},
  year={2026}
}

@inproceedings{peebles2023scalable,
  title={Scalable diffusion models with transformers},
  author={Peebles, William and Xie, Saining},
  booktitle={2023 IEEE/CVF International Conference on Computer Vision (ICCV)},
  pages={4172--4182},
  year={2023},
  organization={IEEE}
}

@inproceedings{
xu2020theory,
title={A Theory of Usable Information under Computational Constraints},
author={Yilun Xu and Shengjia Zhao and Jiaming Song and Russell Stewart and Stefano Ermon},
booktitle={International Conference on Learning Representations},
year={2020},
url={https://openreview.net/forum?id=r1eBeyHFDH}
}

@article{noel2018neural,
  title={Neural adaptation accounts for the dynamic resizing of peripersonal space: evidence from a psychophysical-computational approach},
  author={Noel, Jean-Paul and Blanke, Olaf and Magosso, Elisa and Serino, Andrea},
  journal={Journal of neurophysiology},
  volume={119},
  number={6},
  pages={2307--2333},
  year={2018},
  publisher={American Physiological Society Bethesda, MD}
}

@article{bufacchi2018action,
  title={An action field theory of peripersonal space},
  author={Bufacchi, Rory J and Iannetti, Gian Domenico},
  journal={Trends in cognitive sciences},
  volume={22},
  number={12},
  pages={1076--1090},
  year={2018},
  publisher={Elsevier}
}

@article{kohlbrenner2026egocentric,
  title = {Egocentric Tactile and Proximity Sensors as Observation Priors for Humanoid Collision Avoidance},
  author = {Kohlbrenner, Carson and Pudasaini, Niraj and Xie, William and Sivagnanadasan, Naren and Correll, Nikolaus and Roncone, Alessandro},
  journal = {arXiv preprint arXiv:2604.25554},
  year = {2026},
}
\bibliographystyle{iclr2027_conference}

\appendix
\section{Appendix}

\subsection{Data Curation}
\label{app:data}
The PWM is pretrained on a sample dataset of 6000 trajectories of expert demonstrations containing ray sensor observations and actions. To make sure the PWM can accurately predict contact for wide range of potential policies, one third of the experts were trained having seen only raw observations ($x^\text{raw}$), one third was trained having perfect localization of incoming balls ($f_\text{tp}$), and the remaining third was trained with only the minimum sensor distance ($f_\text{min}$). To ensure that the PWM could predict incoming projectiles during training and not just when the robot has full range of motion, 50\% of the data contains intermediate training step demonstrations, comprised equally of the aforementioned experts at 500, 1000, 1500, 2000, and 2500 steps of training. The remaining 50\% of the data was collected using the fully trained experts at 3000 steps. 15\% of the dataset was used for validation, and an additional hold-out dataset of 1000 trajectories was collected using two experts that were not used for collecting the training dataset to verify that the PWM did not overfit.

The dynamics model learns the transition dynamics between time steps and can fail when the frame rate of the data is too high to capture local dynamics. To prevent this failure mode, the $\qty{60}{Hz}$ dataset was windowed in a $\qty{10}{Hz}$ stride, forcing the dynamics model to predict in $\qty{100}{ms}$ intervals.

Another failure mode of the PWM is too sparse of contact labels in the data to drive effective gradient passes, but by adding contact padding frames after an initial contact occurred, the windows could capture the sparse moments leading up to contact in the full $C+R$ sized window.  

\subsection{World Model Architecture}
\subsubsection{Overview}
\label{app:wm}


\paragraph{Encoder/Decoder Model:} The encoder and decoder used in this work are three-dimensional variational autoencoders (3D-VAE), which stride over the $H$, $W$, and $S$ \citep{kingma2013auto}. A handful of variations to the standard evidence lower bound (ELBO) loss are employed in our approach to improve the fidelity of depth signals, such as a $\beta$ regime to reduce pixel smoothing on the posterior \citep{higgins2017beta}, a deterministic reconstruction term to improve decoder accuracy \citep{ghosh2019variational}, and a sigmoid decoder trained using weighted soft-label binary cross-entropy (\cref{app:wm}).

\paragraph{Dynamics Model:} The dynamics model uses conditional flow matching \citep{lipman2022flow} to generate a sequence of $R$ likely future latent observations $z_{t+1:t+R}$ conditioned on $C$ prior encoded observations, and candidate future actions $\hat{a}$ as
\begin{equation}
    \hat{z}_{t+1:t+R}
\sim
r_\phi(
z_{t+1:t+R}
\mid
z_{t-C+1:t},a_{t}
).
\end{equation}
 The dynamics model draws the full observation roll-out in a single step, making it not auto-regressive to reduce inference time during training. Since the dynamics model is not auto-regressive, we assume a zero-order hold on candidate actions based on the most recent action $a_t$. 



\paragraph{Contact Anticipation:} We attach a causal contact-anticipation head that maps the world model's imagined rollout to a first-contact stopping decision. It takes the fully-denoised context latents plus the jointly-generated rollout latents, pools each frame's sensor tokens to a single frame token, and processes the resulting context+rollout sequence with a causal transformer, yielding one hazard probability $h_r \in [0,1]$ per rollout step where $h_r$ sees only history through $r-1$. At inference we threshold the logits under a sigmoid filter and select the first step above the threshold to define the first contact state in the rollout.

We train the contact anticipation head as a discrete-time first-contact hazard problem rather than per-frame classification: only pre-event steps contribute survival loss $-\log(1-h_t)$, only the first contact step contributes event loss $-\log(h_k)$, and post-contact steps are masked, with an additional margin ranking term forcing the event logit above the hardest pre-event logit in the same window. Training is done on cached world-model generations rather than ground-truth latents, so the head learns the generator's artifacts directly.

The proximity world model is held fixed during policy learning. A frame is a stack of $S=64$ depth images, $x_t\in[0,1]^{1\times 8\times 8\times 64}$, with millimeter ranges divided by $4000$ so that $1$ is the $4\,\mathrm{m}$ full scale. The model uses $C=3$ context frames and predicts $R=9$ future frames. At the $10\,\mathrm{Hz}$ training rate in Appendix~\ref{app:data}, the context covers the preceding $0.2\,\mathrm{s}$ and the rollout covers the next $0.9\,\mathrm{s}$. The three modules are trained in the frozen order of Appendix~\ref{app:train}. Parameter counts are $2.77\times 10^{7}$ for the encoder--decoder, $2.20\times 10^{6}$ for the dynamics model, and $2.86\times 10^{4}$ for the contact head.

\subsubsection{Implementation}

\paragraph{Encoder:} The encoder is a 3D convolutional VAE \citep{kingma2013auto}. Two convolutions with kernel $4$ and stride $(2,2,2)$ map $1\to 256\to 512$ channels and downsample every spatial axis by $4$, followed by a stride-$1$ convolution that preserves $512$ channels. A residual block with hidden width $128$ (kernel-$3$ convolution, ReLU, pointwise convolution, residual add) is then applied eight times with tied weights. Separate $1\times 1\times 1$ convolutions read out the posterior mean and log-variance, with $\log\sigma^2$ clamped to $[-10,10]$:
\begin{equation}
q_\psi(z_t\mid x_t)=\mathcal{N}\!\big(\mu_\psi(x_t),\,\mathrm{diag}\,\sigma^2_\psi(x_t)\big),
\qquad
\mu_\psi(x_t)\in\mathbb{R}^{8\times 2\times 2\times 16}.
\end{equation}
Flattening the spatial axes gives $64$ tokens of dimension $e=8$. The decoder mirrors the encoder with transposed convolutions and ends in a sigmoid, so each depth pixel is a Bernoulli probability. Downstream models consume $\mu_\psi(x_t)$ with no reparameterization noise. Sampling is used only for the stochastic reconstruction term during VAE training.

Depth images are bimodal: most pixels sit at full scale, and the ball occupies a few low values. Mean-squared error therefore blurs the near field. The reconstruction term is a weighted binary cross-entropy,
\begin{equation}
\mathcal{L}_{\mathrm{BCE}}(\hat x,x)
=\mathbb{E}\big[w(x)\,\mathrm{BCE}(\hat x,x)\big],
\qquad
w(x)=
\begin{cases}
10 & x<0.4,\\
1 & x\ge 0.4,
\end{cases}
\end{equation}
so returns closer than $1.6\,\mathrm{m}$ receive $10\times$ reconstruction weight. An $\ell_1$ penalty on horizontal and vertical finite differences, applied inside each sensor and not across sensors, keeps range edges:
\begin{equation}
\mathcal{G}(\hat x,x)
=\|\nabla_H\hat x-\nabla_H x\|_1+\|\nabla_W\hat x-\nabla_W x\|_1.
\end{equation}
Let $\hat x_z$ be the decode of a reparameterized sample and $\hat x_\mu$ the decode of $\mu_\psi(x)$. The training objective is
\begin{align}
\mathcal{L}_{\mathrm{VAE}}
&=
\big[\mathcal{L}_{\mathrm{BCE}}(\hat x_z,x)+\lambda_\mu\mathcal{L}_{\mathrm{BCE}}(\hat x_\mu,x)\big] \nonumber\\
&\quad
+\lambda_\nabla\big[\mathcal{G}(\hat x_z,x)+\lambda_\mu\mathcal{G}(\hat x_\mu,x)\big]
+\beta\,D_{\mathrm{KL}}\!\big(q_\psi(z\mid x)\,\|\,\mathcal{N}(0,I)\big),
\end{align}
with $\lambda_\mu=1.5$, $\lambda_\nabla=0.1$, and $\beta=10^{-4}$. The KL is the mean per-dimension closed form, so $\beta$ is a light regularizer: reconstruction, including the deterministic decode of $\mu$ \citep{ghosh2019variational}, determines the latent, while a nonzero KL keeps dynamics samples in a region the decoder has seen \citep{higgins2017beta}. The encoder is optimized with Adam at learning rate $10^{-4}$ for $100$ epochs and batch size $512$, on frames shuffled independently of time.

\paragraph{Dynamics:} Future latents are drawn jointly with conditional flow matching \citep{lipman2022flow}. For a target window $z_1=z_{t+1:t+R}$ and $z_0\sim\mathcal{N}(0,I)$,
\begin{equation}
z_\tau=(1-\tau)z_0+\tau z_1,
\qquad
\tau\sim\mathrm{Unif}(0,1),
\qquad
\mathcal{L}_{\mathrm{FM}}
=\big\|v_\phi(z_\tau,\tau\mid z_{t-C+1:t},a_t)-(z_1-z_0)\big\|_2^2.
\label{eq:fm}
\end{equation}
The conditional velocity $v_\phi$ is a two-layer diffusion transformer (DiT) of width $256$, four heads, and feed-forward width $512$, with dropout $0.001$ \citep{peebles2023scalable}. Context tokens and noisy future tokens are concatenated. Each token is the sum of a linear map of its $8$-dimensional latent, a learned frame index, a learned index among the $64$ spatial tokens, and a learned role embedding that marks context versus future. The only action input is the $21$-dimensional joint target $a_t$ at the last context frame. A two-layer MLP ($21\to 256\to 256$, SiLU) maps $a_t$ and adds it to every spatial token of every future frame, so the same action is held for all $R$ steps in both training and sampling. Flow time $\tau$ is a sinusoidal embedding that produces adaptive layer-norm scale, shift, and residual gates inside each block, zero-initialized so each block starts as the identity \citep{peebles2023scalable}. Attention is bidirectional over the window. Only the future-token features are projected back to an $8$-dimensional velocity.

\begin{figure}[h]
    \centering
    \includegraphics[width=\linewidth]{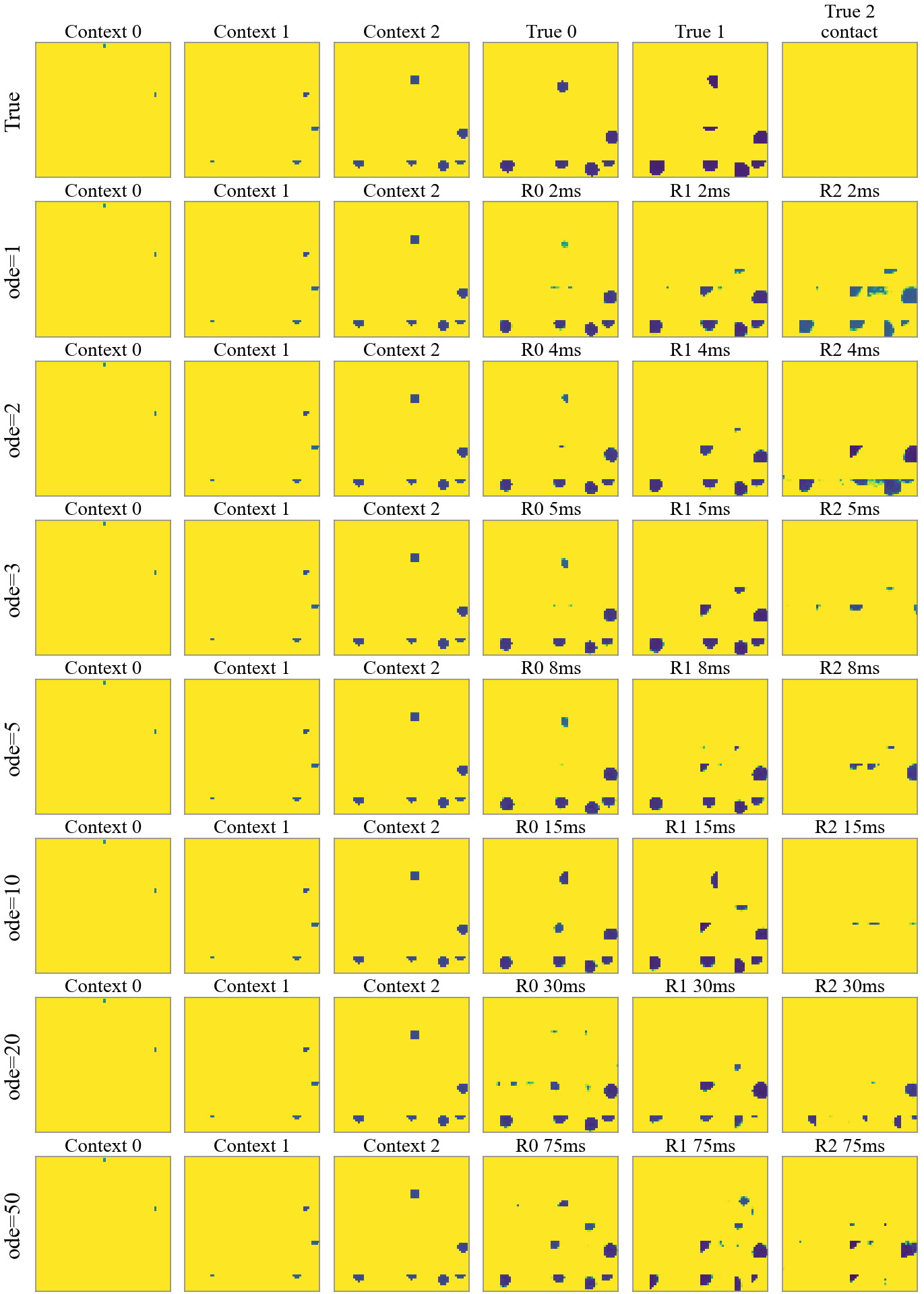}
    \caption{A comparison of Euler step sizes and computation for the dynamics model reveals that only one ODE step is sufficient for reconstruction.}
    \label{fig:ode_sweep}
\end{figure}

Sampling integrates $\mathrm{d}z/\mathrm{d}\tau=v_\phi$ with Euler's method for $3$ uniform steps from $\tau=0$ to $\tau=1$, starting at Gaussian noise. $3$ steps was found to be a sweet spot between computation and prediction accuracy for our application (\cref{fig:ode_sweep}). All $R$ frames are produced by that one integration. The dynamics model is optimized with AdamW at learning rate $5\times 10^{-4}$, weight decay $10^{-3}$, and gradient clip $1$, for $100$ epochs and batch size $512$, on posterior means from the frozen encoder. 

\begin{figure}
    \centering
    \includegraphics[width=0.8\linewidth]{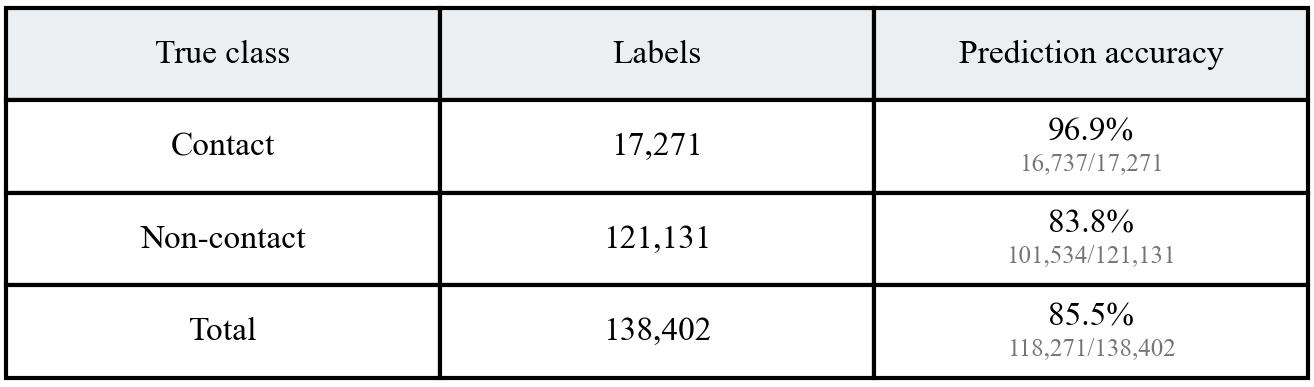}
    \caption{Contact prediction accuracy on the validation dataset.}
    \label{fig:contact_accuracy}
\end{figure}

\paragraph{Contact prediction head:} The head maps a fully denoised context and a generated rollout to a first-contact decision. Within each frame, tokens are projected to width $32$, the same frame, spatial, and role embeddings are added, and a learned query cross-attends to the $64$ tokens. A residual MLP of width $64$ yields one frame vector. A two-layer causal transformer (width $32$, two heads, feed-forward width $64$, dropout $0.05$) then runs over the $C+R$ frame vectors. The hazard logit $\ell_r$ for future step $r\in\{1,\ldots,R\}$ is read from the state that has attended to the context and to generated frames $1,\ldots,r-1$. The frame being scored is excluded, so a generated latent cannot explain its own contact label. The linear head is bias-initialized to $-1.4$, i.e. a prior hazard probability $\sigma(-1.4)\approx 0.20$.

Labels come from the simulator contact flag and may stay positive after impact. Training uses only the first positive step. If that step is $T_c$, steps $t<T_c$ are survival steps, step $T_c$ is the event, and steps $t>T_c$ are masked. A window with no contact contributes only survival terms. With $h=\sigma(\ell)$,
\begin{align}
\mathcal{L}_{\mathrm{event}}
&=\mathbb{E}_{t=k}\big[-\log h_t\big],
&
\mathcal{L}_{\mathrm{surv}}
&=\mathbb{E}_{t<k}\big[-\log(1-h_t)\big],
\label{eq:hazard}\\
\mathcal{L}_{\mathrm{rank}}
&=\mathbb{E}\Big[\mathrm{softplus}\big(\max_{t<k}\ell_t-\ell_k+m\big)\Big],
&
\mathcal{L}_{\mathrm{hazard}}
&=
\mathcal{L}_{\mathrm{event}}+\lambda_s\mathcal{L}_{\mathrm{surv}}+\lambda_r\mathcal{L}_{\mathrm{rank}}.
\nonumber
\end{align}
Averages are taken over the positions or windows that contribute to each term, so the single event step is not diluted by the longer survival prefix. Windows whose contact falls on the first rollout step have no pre-event logit and skip the ranking term. We use $\lambda_s=0.8$, $\lambda_r=0.3$, and $m=0$. The head is optimized with Adam at learning rate $3\times 10^{-4}$ for $200$ epochs and batch size $1024$, on rollouts cached from the frozen dynamics model, so it learns that generator's errors.

At inference, $h_r=\sigma(\ell_r)$ and the selected future frame is
\begin{equation}
r^\star=\min\{r:h_r\ge 0.5\},
\end{equation}
or $r^\star=R$ when no step crosses $0.5$. The policy receives the latent of that frame.

The contact head achieved a 96.9\% prediction accuracy on future contact, and an 83.8\% accuracy on non-contact labels (\cref{fig:contact_accuracy}).


\subsection{Training Regime}
\label{app:train}
The encoder and decoder were trained together without the dynamics and contact prediction models, then held frozen while the dynamics model and contact anticipation head were trained. The dynamics model was also held frozen during training of the contact prediction head.

The encoder and decoder were trained on a shuffled set of raw sensor observations. The dynamics model was trained on temporally consistent latent observations generated by the encoder model, and the contact anticipation head was trained on rollouts of latent vectors generated by the dynamics model.

\paragraph{Limitations:} We note that a large dataset of trajectories was needed to train the PWM itself, rendering it not sample-efficient for adaptive end-to-end collision avoidance on real data. However, we maintain the stance that our investigation reveals \textit{how} representing future sensor observations can influence downstream sample-efficiency during training. We hope that this work advances egocentric proximity sensors do be designed for sample-efficiency downstream and end-to-end on real systems.

\end{document}